\documentclass[conference]{IEEEtran}
\IEEEoverridecommandlockouts
\usepackage{cite}
\usepackage{amsmath,amssymb,amsfonts}
\usepackage{algorithmic}
\usepackage{graphicx}
\usepackage{textcomp}
\usepackage{xcolor}
\usepackage{listings}
\usepackage{booktabs}
\usepackage{multirow}
\usepackage{subfigure}

\usepackage{arydshln}
\usepackage{multirow}
\usepackage{multicol}
\usepackage{bm}
\usepackage{url}
\usepackage{underscore}
\usepackage{color}
\usepackage{colortbl}
\usepackage{makecell}
\usepackage{enumitem}
\usepackage{soul}
\usepackage{arydshln}

\usepackage{xcolor}

\def\BibTeX{{\rm B\kern-.05em{\sc i\kern-.025em b}\kern-.08em
    T\kern-.1667em\lower.7ex\hbox{E}\kern-.125emX}}
\begin{document}

\title{Modality-Aware Negative Sampling for Multi-modal Knowledge Graph Embedding\\
}

\author{\IEEEauthorblockN{\normalsize Yichi Zhang, Mingyang Chen, Wen Zhang$^\dagger$}
\IEEEauthorblockA{
{\normalsize \textit{Zhejiang University}, 
Hangzhou, China} \\
{\small \{zhangyichi2022, mingyangchen, zhang.wen\}@zju.edu.cn} 
}

}
\maketitle
\def\thefootnote{$\dagger$}\footnotetext{Corresponding Author.}

\begin{abstract}
Negative sampling (NS) is widely used in knowledge graph embedding (KGE), which aims to generate negative triples to make a positive-negative contrast during training. However, existing NS methods are unsuitable when multi-modal information is considered in KGE models. They are also inefficient due to their complex design. In this paper, we propose Modality-Aware Negative Sampling (MANS) for multi-modal knowledge graph embedding (MMKGE) to address the mentioned problems. MANS could align structural and visual embeddings for entities in KGs and learn meaningful embeddings to perform better in multi-modal KGE while keeping lightweight and efficient. Empirical results on two benchmarks demonstrate that MANS outperforms existing NS methods. Meanwhile, we make further explorations about MANS to confirm its effectiveness.
\end{abstract}


\section{Introduction}
Knowledge graphs (KGs) \cite{DBLP:conf/semweb/AuerBKLCI07, DBLP:conf/sigmod/BollackerEPST08} represent real-world knowledge in the form of triple $(h,r,t)$, which indicates the entity $h$ and the entity $t$ are connected by the relation $r$.
Multi-modal KGs (MMKGs) are the KGs that consist of rich modal information such as images and text. Nowadays, KGs and MMKGs have been widely used in AI-related tasks like question answering \cite{DBLP:conf/naacl/YasunagaRBLL21}, recommendation systems \cite{DBLP:conf/icde/ZhangWYWZC21}, language modeling \cite{DBLP:conf/aaai/LiuZ0WJD020} and telecom fault analysis \cite{chen2022tele}.
\par Meanwhile, KGs as well as MMKGs are usually far from complete and comprehensive because many triples are unobserved, which restricts the application of KGs and makes 
knowledge graph completion (KGC)
a significant task. Knowledge graph embedding (KGE) \cite{DBLP:conf/nips/BordesUGWY13,DBLP:journals/jmlr/TrouillonDGWRB17,DBLP:journals/corr/YangYHGD14a, DBLP:conf/iclr/SunDNT19} is a popular and universal approach for KGC, which represents entities and relations of KGs in a continuous low-dimension vector space. In the usual paradigm, KGE models would design a score function to estimate the plausibility of triples with entity and relation embeddings. These embeddings are structural embeddings since they can encode information about triple structures. As for MMKGs, embedding-based methods can still work by utilizing multi-modal information. Nevertheless, existing multi-modal KGE (MMKGE) \cite{DBLP:conf/ijcai/XieLLS17, DBLP:conf/emnlp/PezeshkpourC018, DBLP:conf/ijcnn/WangLLZ19} methods design additional embeddings to represent the modal information, which would also participate in the score function.
\par Negative sampling (NS) \cite{DBLP:conf/nips/BordesUGWY13} is a widely used technology for training KGE models, which aims to generate manual negative triples by randomly replacing entities for positive-negative contrast. NS would guide the KGE model to give higher scores for the positive triples. An outstanding NS strategy would obviously improve the performance of KGE models to discriminate the triple plausibility.

\par Though existing NS methods \cite{DBLP:conf/aaai/WangLP18, DBLP:conf/icde/ZhangYSC19, DBLP:conf/emnlp/AhrabianFSHB20, DBLP:conf/www/LiJFGXCZ21, DBLP:conf/acl/Niu0ZP22, DBLP:journals/corr/abs-2210-06242} have tried different ways to obtain high-quality negative samples, they have one drawback that cannot be ignored: they are designed for general KGE models and \textbf{underperform in MMKGE}. As for MMKGE, entities may have multiple heterogeneous embeddings such as visual and structural embeddings. However, NS for the general KGE models will treat multiple embeddings of an entity as a whole and replace them together with embeddings of another entity, which we think is entity-level. Such design implicitly assumes that different embeddings of an entity have been aligned and model could distinguish the two embeddings of each entity, which weakens the model's capability of aligning different embeddings and results in less semantic information being learned by the embeddings.
Besides, we should also take the efficiency of the method into account while considering the multi-modal scenario, as those existing approaches design many complex modules (e.g. GAN \cite{DBLP:conf/aaai/WangLP18}, large-scale caches \cite{DBLP:conf/icde/ZhangYSC19}, manual rules \cite{DBLP:conf/acl/Niu0ZP22}, entity clustering \cite{DBLP:journals/corr/abs-2210-06242}) to sample high-quality negative samples. We think they are over-designed and make the NS method computationally expensive.

\par To address the mentioned challenges, we propose \textbf{M}odality-\textbf{A}ware \textbf{N}egative \textbf{S}ampling (MANS for short) strategy for MMKGE. MANS is a lightweight but effective NS strategy designed for MMKGE. We first propose visual NS (MANS-V for short), a modal-level sampling strategy that would sample only negative visual features for contrast. We employ MANS-V to achieve modality alignment for multiple entity embeddings and guide the model to learn more semantic information from different perspectives by utilizing multi-modal information. We further extend MANS-V to three combined strategies, called two-stage, hybrid, and adaptive negative sampling respectively. All of the NS methods make up MANS together. Our Contribution could be summarized as follows:
\begin{itemize}
    \item To the best of our knowledge, MANS is the first work focusing on the negative sampling strategy for multi-modal knowledge graph embedding.
    \item In MANS, we propose MANS-V to align different modal information. Furthermore, we extend it to three combined NS strategies with different settings.
    \item We conduct comprehensive experiments on two knowledge graph completion tasks with two MMKG datasets. Experiment results illustrate that MANS could outperform the baseline methods in various tasks. 
    \item We further carry out extensive analysis to explore several research questions about MANS to demonstrate the details of MANS.
\end{itemize}

\section{Related Works}
\subsection{Knowledge Graph Embedding}
Knowledge Graph Embedding (KGE) \cite{DBLP:journals/tkde/WangMWG17} is an important research topic for knowledge graphs, which focuses on embedding the entities and relations of KGs into low-dimensional continuous vector space. 
\par General KGE methods utilize the triple structure to embed entities and relations and follow the research paradigm that defines a score function to measure the plausibility of triples in the given KG. Negative sampling (NS)
is a significant technology widely used when training KGE models. During training, positive triples should get higher scores than those negative triples, which are generated by NS. 
\par Previous KGE methods can be cursorily divided into several categories. Translation-based methods like TransE \cite{DBLP:conf/nips/BordesUGWY13} and TransH \cite{DBLP:conf/aaai/WangZFC14} modeling the triples as the translation from head to tail entities with a distance-based scoring function. Semantic-based methods like DistMult \cite{DBLP:journals/corr/YangYHGD14a} and ComplEx \cite{DBLP:journals/jmlr/TrouillonDGWRB17} use similarity-based scoring functions. Neural network-based methods \cite{DBLP:conf/aaai/DettmersMS018, DBLP:conf/naacl/JiangWW19} employ neural networks to capture features from entities and relations and score the triples. Several KGE methods modeling triples with various mathematical structures, such as RotatE \cite{DBLP:conf/iclr/SunDNT19}, ConE \cite{DBLP:conf/nips/ZhangWCJW21}.  Some recent methods \cite{DBLP:conf/www/ZhangPWCZZBC19, zhen2023analogical} combine rule learning / analogical inference and KGE together to enhance the interpretability of KGE models.

\subsection{Multi-modal Knowledge Graph Embedding}
The KGE methods mentioned before are unimodal approaches as they only utilize the structure information from KGs. For multi-modal Knowledge Graphs (MMKGs), the modal information like images and text should also be highly concerned as another embedding for each entity and relation.
Existing methods usually extract modal information using pre-trained models and project the modal information into the same representation space as structural information. IKRL \cite{DBLP:conf/ijcai/XieLLS17} apply VGG  \cite{DBLP:journals/corr/SimonyanZ14a} to extract visual information of entities' images and scoring a triple with both visual information and structure information using TransE \cite{DBLP:conf/nips/BordesUGWY13}. TransAE \cite{DBLP:conf/ijcnn/WangLLZ19} also employs TransE as the score function and exact modal information with a multi-modal auto-encoder. Mosselly et al  \cite{DBLP:conf/starsem/SergiehBGR18} and Pezeshkpour et al  \cite{DBLP:conf/emnlp/PezeshkpourC018} use VGG \cite{DBLP:journals/corr/SimonyanZ14a}
 and GloVe \cite{DBLP:conf/emnlp/PenningtonSM14} to separately extract visual and textual information and then fused them into multi-modal information. Recently, RSME \cite{DBLP:conf/mm/WangWYZCQ21} focused on preserving truly valuable images and discarding the useless ones with three gates.

\subsection{Negative Sampling in Knowledge Graph Embedding}

Negative sampling (NS) aims to generate negative triples which don't appear in existing KGs. Those negative triples will participate in the training process of KGE models by contrasting them with positive triples. Therefore, many NS methods are proposed to generate high-quality negative samples. Normal NS \cite{DBLP:conf/nips/BordesUGWY13} randomly replaces the head or tail entity with another entity with the same probabilities. KBGAN \cite{DBLP:conf/naacl/CaiW18} and IGAN \cite{DBLP:conf/aaai/WangLP18} apply Generative Adversarial Networks (GANs) \cite{DBLP:conf/nips/GoodfellowPMXWOCB14} to select harder negative samples. NSCaching \cite{DBLP:conf/icde/ZhangYSC19} store the high-quality negative triples with cache during training to achieve efficient sampling. NS-KGE \cite{DBLP:conf/www/LiJFGXCZ21} employs a unified square loss to avoid NS during training. It is called no-sampling but all-sampling. SANS \cite{DBLP:conf/emnlp/AhrabianFSHB20} utilize the graph structure to sample high-quality negative samples. CAKE \cite{DBLP:conf/acl/Niu0ZP22} construct commonsense from KGs to guide NS. EANS \cite{DBLP:journals/corr/abs-2210-06242} propose a clustering-based negative sampling strategy with an auxiliary loss function. VBKGC \cite{zhang2022knowledge} propose a twins negative sampling method for different parts of the score function.
\par However, many of the NS methods have their shortcomings which leads to the dilemma of NS for MMKGE. On the one hand, they are not lightweight enough as extra modules are introduced in the models. On the other hand, they are designed for unimodal knowledge graph embedding. 
Such a strategy performs well in general KGE because each entity has only one structural embedding. As many MMKGE models define multiple embeddings for each entity, the alignment between different embeddings is also significant but ignored by existing methods.

\section{Problem Formulation}
In this section, we would introduce the basic pipeline of multi-modal knowledge graph embedding (MMKGE) in a three-step format. We first formally describe what a MMKG is and the embeddings we design for the MMKGE task. Then we detailedly introduce the modules of the MMKGE model. Eventually, we would show the training  objective of MMKGE model and emphasis the process of negative sampling.

\subsection{Basic Definition}
A MMKG can be denoted as $
\mathcal{G}_{M}=(\mathcal{E},\mathcal{R},\mathcal{I},\mathcal{T})
$, where $\mathcal{E},\mathcal{R},\mathcal{I},\mathcal{T}$ 
are the entity set, relation set, image set and triple set. Entities in $\mathcal{E}$ may have 0 to any number of images in $\mathcal{I}$, and the image set of entity $
e$ is denoted as $
I_e$. 
\par We denote $
\mathbf{e}_s$ and $\mathbf{e}_v$ as the structural embedding and visual embedding for an entity $
e$, respectively. Therefore, the entity $
e$  can be represented by two embedding vectors $\mathbf{e}_s, \mathbf{e}_v$. Besides, we denote $\mathbf{r}$ as the structural embedding of relation $r$.

\subsection{MMKGE Framework}

\label{sec:framework}
\begin{figure}[h]
  \centering
  \includegraphics[width=\linewidth]{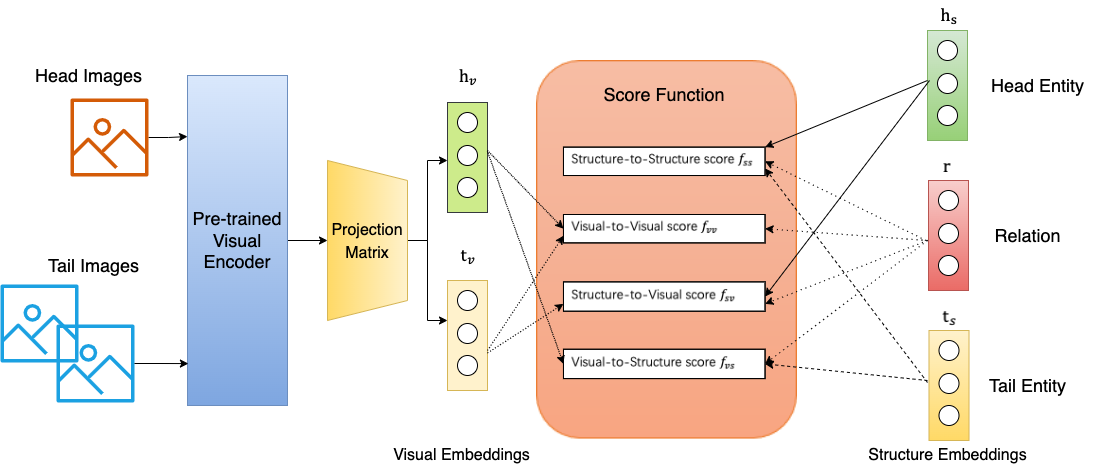}
  \caption{Our Multi-modal KGE model architecture}
  \label{img:model}
\end{figure}

In this paper, we employ a general MMKGE framework as the backbone model. The model architecture is shown in Figure \ref{img:model}, which consists of a visual encoder and a score function.

\subsubsection{Visual Encoder}

\par Visual encoder, which is denoted as $E_
{img}$, aims to capture the visual feature of entities and project them into the same representation space of structural embeddings.
For those entities with more than one image, we use mean pooling to aggregate the visual feature. The visual embedding $\mathbf{e}_v$ of entity $e$ can be denoted as:
\begin{equation}
    \mathbf{e}_v=\mathbf{W} \times \frac{1}{|I_e|}  \sum_{I_{e}^k \in I_e }E_{img}(I_e^k)
\end{equation}
where  $\mathbf{W} \in \mathbb{R}^{d \times d_v}$ is the projection matrix,  $
d$ is the dimension of both structural and visual embedding and $
d_v$ is the dimension of the output dimension of the visual encoder. In this paper, we employ pre-trained VGG-16 \cite{DBLP:journals/corr/SimonyanZ14a} as the visual encoder.
\subsubsection{Score Function}

The score function is denoted as $\mathcal{F}(h, r, t)$. Both the structural embeddings $\mathbf{e}_s$ and visual embeddings $\mathbf{e}_v$ will be considered in the score function. The overall score function consists of four parts, aiming to learn the embeddings in the same vector space, which can be denoted as:
$
    \mathcal{F}(h, r, t)=f(\mathbf{h_s},\mathbf{r},\mathbf{t_s})+f(\mathbf{h_v},\mathbf{r},\mathbf{t_v})+f(\mathbf{h_s},\mathbf{r},\mathbf{t_v})\\
    +f(\mathbf{h_v},\mathbf{r},\mathbf{t_s})
$, 
where $f$ is the TransE score \cite{DBLP:conf/nips/BordesUGWY13}.

\par Besides, the overall score function $\mathcal{F}(h, r, t)$ can be divided into two parts, unimodal scores, and multi-modal scores. The unimodal scores only consider single-modal embedding of entities while multi-modal scores use both structural embeddings and visual embeddings. Under such criteria, $f(\mathbf{h_s},\mathbf{r},\mathbf{t_s}), f(\mathbf{h_v},\mathbf{r},\mathbf{t_v})$ are unimodal scores and $f(\mathbf{h_s},\mathbf{r},\mathbf{t_v}), f(\mathbf{h_v},\mathbf{r},\mathbf{t_s})$ are multi-modal scores. Such a distinction of scores will play an important role in adaptive NS.
\subsection{Sampling and Training}
The general target of a MMKGE model is to give higher scores for the positive triples and lower scores for the negative triples. In another word, the MMKGE model would discriminate the plausibility of a given triple by its score, which is widely used in KGC to predict the missing triples. Margin-rank loss is a general training objective extensively used in the MMKGE model \cite{DBLP:conf/ijcai/XieLLS17, DBLP:conf/emnlp/PezeshkpourC018}. It could be denoted as:
\begin{equation}
    \mathcal{L}=\sum_{(h, r, t) \in \mathcal{T}}\sum_{(h', r', t')\in \mathcal{T}'}\max (\gamma-\mathcal{F}(h, r, t)+\mathcal{F}(h', r', t'))
\end{equation}
where $\gamma$ is the margin, $(h, r, t)$ is the positve triple in the KG and $(h', r', t')$ is the negative triples.
\par Besides, a given KG usually consists of the observed facts, which are all positive triples. We need to generate the negative triple $(h',r',t')$ manually. Such a process is what we call negative sampling (NS). In normal NS, one of the head and tail entities is randomly replaced. In this setting, $h', t'$ are still the entities in $\mathcal{E}$. This also means that normal NS is an entity-level sampling strategy as it samples negative entities for a given positive triple. As we have analyzed in the previous section, normal NS is suitable for general KGE models but fails when it comes to the MMKGE. In the next section, we will introduce our NS methods to sample better negative triples.

\section{Methodology}
\label{sec:negative}
Normal NS is an entity-level strategy, as all the embeddings of the selected entity are replaced by the negative ones. However, our approach differs. In this section, we would briefly introduce our \textbf{M}odality-\textbf{A}ware \textbf{N}egative \textbf{S}ampling (MANS). MANS is based on visual negative sampling (MANS-V for short), which is a modal-level NS strategy and would sample negative visual embeddings for a finer contrast. We further combine MANS-V and normal NS with a sampling proportion $\beta$ and propose three more comprehensive NS settings. They are two-stage negative sampling (MANS-T), hybrid negative sampling (MANS-H), and adaptive negative sampling (MANS-A).
\subsection{Visual Negative Sampling (MANS-V)}
MANS-V aims to sample the negative visual embeddings that do not belong to the current entity to teach the model to identify the visual features corresponding to each entity, which could achieve the modality alignment between structural and visual embeddings. In our context, modality alignment means that the model could identify the relations between the two modal embeddings, which we think is of great importance in MMKGE.
\par
\begin{figure}
  \centering
  \includegraphics[width=\linewidth]{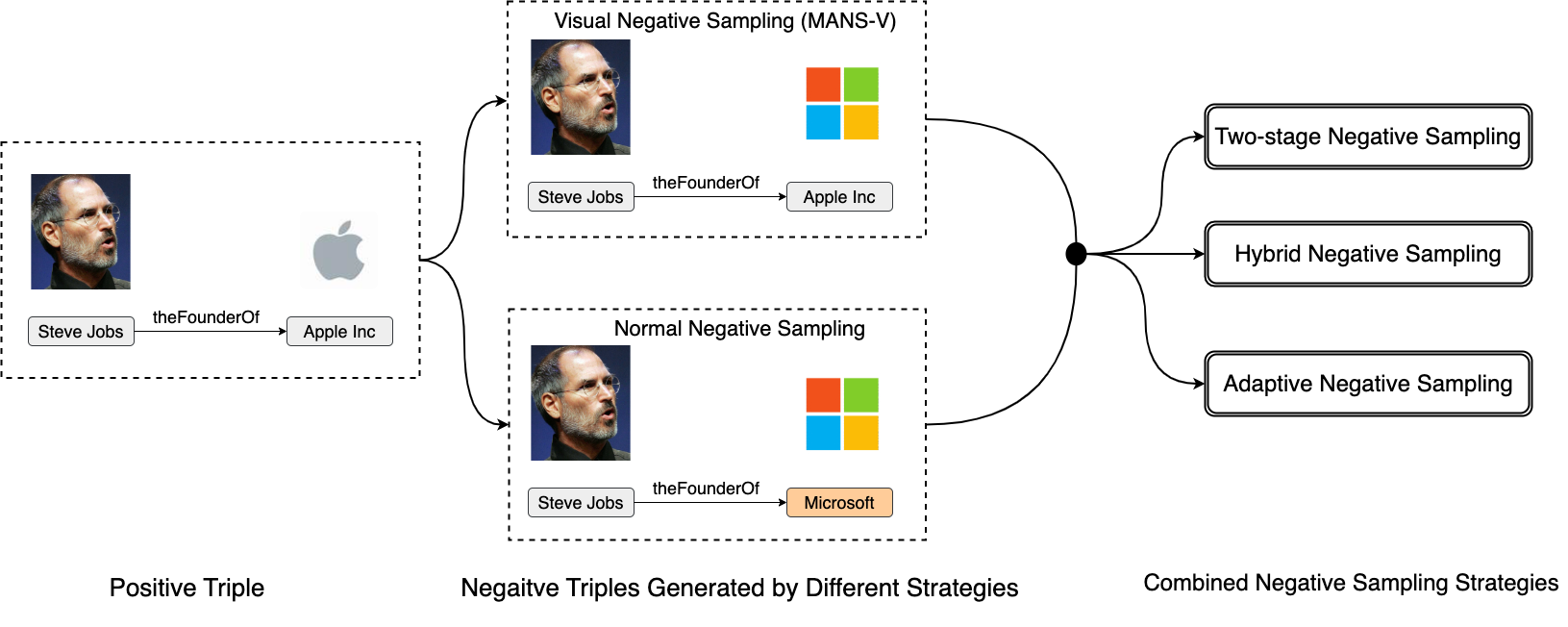}
  \caption{An example of MANS-V. Only negative visual feature is sampled compared with normal negative sampling. We further combine MANS-V with normal NS to get three more NS strategies.}
  \label{img:sampling}
\end{figure}

 MANS-V is a modal-level method that would sample negative visual embeddings. The negative triple $(h',r',t')$ generated by MANS-V preserves the original structural embeddings but the visual embedding of the replaced entity is changed. For example, if we replace head entity $h$ with another entity $h'$, the embeddings of $h'$ used during training is $\mathbf{h}_s, \mathbf{h}'_v$. For tail entity, the embeddings of $t'$ is  $\mathbf{t}_s, \mathbf{t}'_v$. In MANS-V, the replaced entity is a virtual negative entity that doesn't exist in $\mathcal{E}$. An intuitive example of MANS-V is shown in Figure \ref{img:sampling}.
\par Thus, MANS-V is a more fine-grained strategy compared with normal sampling. It changes the granularity of NS from the whole entity to the single modal embedding of the entity. By sampling only negative visual embeddings, MANS-V could achieve alignment between different modal embeddings for an entity.
\par KGE models would learn to align the two embeddings for each entity by MANS-V. However, learning to discriminate the plausibility of triples is still significant, which could be achieved by normal NS. Hence, we consider that MANS-V could play an important role as the auxiliary to enhance the normal NS and we propose three combination strategies for comprehensive training.
\subsection{Two-Stage Negative Sampling (MANS-T)}
MANS-T divides the training process into two different stages:
\begin{itemize}
    \item In \textbf{Stage1}, MANS-V is applied to train the model. The model would learn to align different modal embeddings in this stage.
    \item In \textbf{Stage2}, we employ normal sampling and train the model to discriminate the plausibility. As the structural and visual embeddings are aligned inside each entity, the model would learn better in this stage.
\end{itemize}
\par We assume that the total training epoch is $
M$ and the proportion of MANS-V is $\beta_1$, then the turning point for stage switching is:
\begin{equation}
    M_0=\beta_1 \times M
\end{equation}
which means training epoch $[0, M_0]$ is  \textbf{Stage1}  and $[M_0 + 1, M]$ is \textbf{Stage2}.
It's not difficult to find that normal NS and MANS-V are two special cases of MANS-T when $\beta_1=0$ (normal) or $\beta_1=1$ (image).
\subsection{Hybrid Negative Sampling (MANS-H)}
As MANS-T divides the NS from the view of training epochs, MANS-H would apply two sampling strategies in each training epoch. Compared  with the two-stage setting, MANS-H is more progressive.
\par In each mini-batch of one training epoch, we assume that the batch size is $N
$ and the MANS-V proportion is $\beta_2$, for each triple, we sample $k
$ negative samples, then the total number of negative samples is $kN$ and the total negative triples generated by MANS-V is:
\begin{equation}
    N_0 = \beta_2\times kN
\end{equation}
which means that $N_0$ negative samples are randomly generated by MANS-V in a mini-batch and others are generated by normal NS. During the whole training process, MANS-H will be applied and the negative samples are blended from multiple sampling strategies. In MANS-H, the sampling proportion $\beta_2$ is a tunable hyper-parameter. The same as the two-stage setting, MANS-H becomes normal NS when $\beta_2=0$ and MANS-V when $\beta_2=1$.
\subsection{Adaptive Negative Sampling (MANS-A)}
MANS-A is an improved version of MANS-H, which no longer needs to tune the sampling proportion anymore. MANS-A will change the proportion $\beta_3$ adaptively. The adaptive sampling proportion $\beta_3$ would be determined by different scores of the training data.
\par As mentioned before, the overall score function $\mathcal{F}(h, r, t)$ can be divided into unimodal scores and multi-modal scores. We could denote the two parts as:
\begin{equation}
    \mathcal{F}_{unimodal}(h, r, t)=f(\mathbf{h_s},\mathbf{r},\mathbf{t_s}) + f(\mathbf{h_v},\mathbf{r},\mathbf{t_v})
\end{equation}
\begin{equation}
    \mathcal{F}_{multimodal}(h, r, t)=f(\mathbf{h_s},\mathbf{r},\mathbf{t_v}) + f(\mathbf{h_v},\mathbf{r},\mathbf{t_s})
\end{equation}
We define a function $\Phi(h_i,r_i,t_i)$ to discriminate whether the triple $(h_i,r_i,t_i)$ need MANS-V. The function $\Phi(h_i,r_i,t_i)$ is defined as:
\begin{equation}
   \Phi(h_i,r_i,t_i)=\left\{\begin{array}{ll}
0& \mathcal{F}_{multimodal} \geq \mathcal{F}_{unimodal}\\
1& \mathcal{F}_{multimodal} < \mathcal{F}_{unimodal}

\end{array}\right.
\end{equation}
which means that, when multi-modal score $\mathcal{F}_{multimodal}(h_i, r_i, t_i)$ is higher than the unimodal score, MANS-V will be applied. As MANS-V would align different modal embeddings and achieve higher multi-modal scores.
Hence, the adaptive proportion $\beta_3$ for each batch is defined as:
\begin{equation}
    \beta_3=\frac{1}{N}\sum_{i=1}^{N}\Phi(h_i,r_i,t_i)
\end{equation}
where $(h_i,r_i,t_i) (i=1,2,\dots, N)$ is the batch data. With sampling proportion $\beta_3$, the MANS-H would be applied during the training of this batch. The biggest difference between adaptive and MANS-H is that we define an adaptive sampling proportion $\beta_3$ and no longer need to tune it anymore, which could reduce the workload for searching better hyper-parameters.

\section{Experiments}
In this section, we will present the detailed experiment settings and the experimental results to show the advantages of MANS. We design several experiments to answer the following research questions (RQs):
\begin{itemize}
    \item \textbf{RQ1:} Could MANS outperform the baseline methods and achieve new state-of-the-art (SOTA) results in various KGC tasks?
    \item \textbf{RQ2:} As a new hyper-parameter $\beta$ is introduced in our method,  how to select better sampling proportion $\beta_i (i=1,2)$ for MANS-T and MANS-H?
    \item \textbf{RQ3:} Is MANS-A a reasonable and effective design? What is the trend of the sampling proportion $\beta_3$ in MANS-A during training?
    \item \textbf{RQ4:} Is MANS efficient and lightweight compared with existing NS methods?
    \item \textbf{RQ5:} Could MANS learn better embeddings with more semantic informarion compared with normal NS?
\end{itemize}

\subsection{Datasets}
\begin{table}[!htbp]
\centering
  \caption{Statistics of datasets }
  \label{tab:dataset}
  \centering
  \begin{tabular}{cccccl}
    
    \toprule
    Dataset & Entities & Relations & Images & Triples \\
    \midrule
    FB15K & 14951 & 1345 & 13444 & 592213\\
    DB15K & 14777 & 279 & 12841 & 99028  \\
    \bottomrule
  
  \end{tabular}
\end{table}
In our experiments, we use two well-known MMKG datasets (FB15K, DB15K with extra images of entities) proposed in \cite{DBLP:conf/esws/LiuLGNOR19}, the statistical information of the datasets is shown in Table \ref{tab:dataset}.

\subsection{Evaluation and Implementation Details}
\begin{table*}[]
\centering
\caption{Evaluation results for link prediction. The best results of each metric  are in bold and the second best results are underlined.}
\label{tab:linkprediction}
\centering
\begin{tabular}{c|ccccc|ccccc}
\toprule
\multirow{2}{*}{Model} & \multicolumn{5}{c|}{FB15K}                                                      & \multicolumn{5}{c}{DB15K}                                                        \\  
                       & MRR$\uparrow$           & MR$\downarrow$          & Hit@10$\uparrow$         & Hit@3$\uparrow$          & Hit@1$\uparrow$          & MRR$\uparrow$            & MR$\downarrow$           & Hit@10$\uparrow$         & Hit@3$\uparrow$          & Hit@1$\uparrow$          \\ \midrule
Normal \cite{DBLP:conf/nips/BordesUGWY13}                 & 0.479          & 95          & \underline{0.755}          & 0.604          & 0.314          & 0.303          & 685          & 0.542          & 0.410          & 0.167          \\
No-Samp \cite{DBLP:conf/www/LiJFGXCZ21}                & 0.109          & 1594        & 0.212          & 0.130          & 0.051          & 0.151          & \textbf{456} & 0.271          & 0.171          & 0.087          \\
NSCaching \cite{DBLP:conf/icde/ZhangYSC19}              & 0.329          & 121         & 0.526          & 0.374          & 0.224          & 0.291          & 835          & 0.471          & 0.344          & 0.192          \\
SANS      \cite{DBLP:conf/emnlp/AhrabianFSHB20}             & 0.394          & 109         & 0.635          & 0.466          & 0.264          & 0.276                &    703      &     0.413          &  0.387               &  0.127          \\
CAKE \cite{DBLP:conf/acl/Niu0ZP22}                  & 0.395          & \textbf{68} & 0.647          & 0.467          & 0.262          & -              & -            & -              & -              & -              \\
EANS \cite{DBLP:journals/corr/abs-2210-06242}                   & 0.483          & 111         & 0.739           & 0.597          & 0.327          & 0.269              & 1036            & 0.489            & 0.353             & 0.141              \\ \midrule
MANS-V                 & 0.454          & 103         & 0.713          & 0.552          & 0.305          & 0.274          & 506          & 0.525          & 0.333          & 0.165          \\
MANS-T                & 0.485          & 93          & 0.748          & 0.591          & 0.333          & 0.307          & 615          & \underline{0.546}         & 0.411          & 0.178          \\
MANS-H                 & \underline{0.493}          & 92          & \textbf{0.756} & \textbf{0.606} & \underline{0.351}        & \underline{0.329} & 553          & 0.541          & \underline{0.414} & \underline{0.204} \\
MANS-A                 & \textbf{0.499} & \underline{88}          & 0.749          & \underline{0.601}          & \textbf{0.353} & \textbf{0.332}          & \underline{549} & \textbf{0.550} & \textbf{0.420}          & \textbf{0.204}     \\
\bottomrule
\end{tabular}
\end{table*}

\subsubsection{Tasks and Evaluation Protocol}
We evaluate our method on two tasks, link prediction, and triple classification \cite{DBLP:conf/nips/BordesUGWY13}. The link prediction task aims to predict the missing entity for a given query $(h, r, ?)$ or $(?, r, t)$ with the KGE model. We evaluate the link prediction task by mean rank (MR) \cite{DBLP:conf/nips/BordesUGWY13}, mean reciprocal rank (MRR) \cite{DBLP:conf/iclr/SunDNT19} and Hit@K (K=1,3,10) \cite{DBLP:conf/nips/BordesUGWY13}.  
Besides, we follow the filter setting \cite{DBLP:conf/nips/BordesUGWY13} which would remove candidate triples that have already appeared in the datasets.
\par Triple classification task would predict the given triple $(h, r, t)$ is true or not. Thus, we evaluate the task with accuracy (Acc), precision (P), recall (R), and F1-score (F1), which are the common metrics for the binary classification task.
\subsubsection{Baselines}
For the link prediction task, we employ the normal NS \cite{DBLP:conf/nips/BordesUGWY13} and several recent SOTA NS methods as the baselines. They are No-Samp \cite{DBLP:conf/www/LiJFGXCZ21}, NSCaching \cite{DBLP:conf/icde/ZhangYSC19}, SANS \cite{DBLP:conf/emnlp/AhrabianFSHB20}, CAKE \cite{DBLP:conf/acl/Niu0ZP22}, and EANS \cite{DBLP:journals/corr/abs-2210-06242}, which enhance the normal NS from their different perspectives. We utilize their official code to conduct baseline results. For the triple classification task, we compare the performance of MANS with normal NS, as other NS methods do not focus on this task and give the corresponding implementations.
\subsubsection{Experiments Settings}
For experiments, we set both structural embedding and visual embedding size $d_e=128$ for each model. The dimension of visual features captured by a pre-trained VGG-16 model is $d_v=4096$. For those entities which have no image, we employ Xavier initialization \cite{DBLP:journals/jmlr/GlorotB10} for their visual features. We set the number of negative triples to 1 and train each model with 1000 epochs.
\par During training, we divide each dataset into 400 batches and apply IKRL \cite{DBLP:conf/ijcai/XieLLS17} as the MMKGE model. We use the default Adam optimizer for optimization and tune the hyper-parameters of our model with grid search. The margin $\gamma$ is tuned in $\{4.0, 6.0, 8.0, 12.0\}$ and learning rate $\eta$ is tuned in $\{0.001, 0.01, 0.1, 1\}$. Besides, for two-stage and MANS-H, we tuned the sampling proportion $\beta_1, \beta_2$ from $0.1$ to $1.0$.
\par For baselines, we have taken full account of the parameter settings in the original paper \cite{DBLP:conf/icde/ZhangYSC19, DBLP:conf/www/LiJFGXCZ21, DBLP:conf/acl/Niu0ZP22, DBLP:journals/corr/abs-2210-06242}. All the experiments are conducted on one Nvidia GeForce 3090 GPU. Our code of MANS is released in https://github.com/zjukg/MANS.

\subsection{RQ1: Main Results}
To answer RQ1, we conduct experiments on two KGC tasks.
The evaluation results of the link prediction task are shown in Table \ref{tab:linkprediction} and the triple classification results are in Table \ref{tab:triple}. From the experimental results, We can conclude the following points:\\
\textbf{Poor performance of the baselines.}
We could observe that existing NS methods have poor performance and they are even worse than the normal NS. According to our previous analysis, these NS methods are designed for general KGE models and are unsuitable for the multi-modal scenario where modal information is carefully considered. They could not align different embeddings of each entity and get bad performance in MMKGE. \\
\textbf{The outperformance of MANS.}
MANS could achieve better link prediction results compared with baselines. For example, MANS-A achieves much better Hit@1 on FB15K compared with baselines (from 0.318 to 0.353, a relative improvement of 9.9\%). Besides, MANS performs particularly well in Hit@1 and MRR, which are sensitive to high-rank results \cite{DBLP:conf/icde/ZhangYSC19}. This means that MANS can largely improve the accurate discriminatory ability of the model by aligning structural and visual embeddings.\\
\textbf{Necessity and effectiveness of MANS-V.}  According to the previous section, MANS-V is designed to align different modal information. Though it does not perform better than baseline methods, MANS-V is the fundamental component of the other three settings of MANS. Besides, we could prove with such a result that both modal alignment and positive-negative discrimination are important for MMKGE, which could be achieved by MANS-V and normal NS respectively. MANS-T, MANS-H, and MANS-A could perform better because they combine the advantages of both. In summary, MANS-V is a necessary design for MMKGE. \\
\textbf{Comparison of different MANS settings.} As we propose three different settings of MANS, we could observe from Table \ref{tab:linkprediction} that all of the three settings (MANS-T, MANS-H, MANS-A) outperform the baseline methods. Experiment results demonstrate that MANS-H and MANS-A would perform better than MANS-T. Meanwhile, MANS-H and MANS-A have their advantages on different datasets and metrics, but the overall difference of link prediction performance between MANS-H and MANS-A is not notable. Nevertheless, the proportion $\beta_2$ of MANS-H needs to be tuned several times to find the best choice while MANS-A could adaptively change the proportion $\beta_3$ during training and get good performance without hyper-parameter tuning. For the mentioned reasons, we believe that the overall performance of MANS-A is better than MANS-T and MANS-H. MANS-A is free of proportion tuning and could achieve outstanding results.\\
\textbf{Universality of MANS.} From Table \ref{tab:triple}, we could see that three settings of MANS could achieve better triple classification results on four metrics compared with normal NS. Besides, MANS-A outperforms MANS-T and MANS-H on accuracy and F1-score. In summary, the results show that our design of MANS could benefit the MMKGE model in various KGC tasks such as link prediction and triple classification, which means that MANS is a universal approach for better KGC.

\begin{table}[]
\centering
\caption{evaluation results for triple classification}
\label{tab:triple}
\centering
\begin{tabular}{c|ccccc}

\toprule
Dataset                & Model  & Accuracy           & Precision             & Recall             & F1-score            \\ \midrule
\multirow{4}{*}{FB15K} & Normal & 95.2          & 94.7          & 95.7          & 95.2          \\
                       & MANS-V & 96.5          &   95.8       & 97.2 & 96.5 \\
                       & MANS-T & 96.2          & 95.7         & 96.8 & 96.2          \\
                       & MANS-H & 96.5          & 95.9          & \textbf{97.3} & 96.5          \\
                       & MANS-A & \textbf{96.6} & \textbf{96.1} & 97.2          & \textbf{96.7} \\ \midrule
\multirow{4}{*}{DB15K} & Normal & 86.6          & 88.1          & 84.7          & 86.4          \\
                       & MANS-V &  85.6        & 85.3  &  85.9       & 85.7 \\
                       & MANS-T & 87.4         & 88.1 & 86.4        & 87.3         \\
                       & MANS-H & 87.9          & \textbf{87.3} & 88.9          & 88.1          \\
                       & MANS-A & \textbf{88.0} & 87.1          & \textbf{89.2} & \textbf{88.1} \\
\bottomrule

\end{tabular}
\end{table}

\subsection{RQ2: Proportion Selection}
\begin{table}[h]
\centering
\caption{Best sampling proportion ($\beta_1$ for MANS-T and $\beta_2$ for MANS-H) for link prediction task.}
\label{tab:pro}
\begin{tabular}{ccc}

\toprule
        & FB15K & DB15K \\ \midrule
MANS-T & 0.4   & 0.3   \\
MANS-H  & 0.3   & 0.3  \\
\bottomrule

\end{tabular}
\end{table}
Though MANS achieved good performance on link prediction and other tasks, a fact cannot be ignored is that MANS might require more effort to tune the sampling proportion ($\beta_1, \beta_2$ for MANS-T and MANS-H respectively). The optimal proportions for MANS-T and MANS-H are shown in Table \ref{tab:pro}, we further explore the impact of sampling proportion on the link prediction task. It would answer RQ2 and guide us in choosing the best sampling proportion.

\begin{figure}[]
  \centering
  \includegraphics[width=0.6\linewidth]{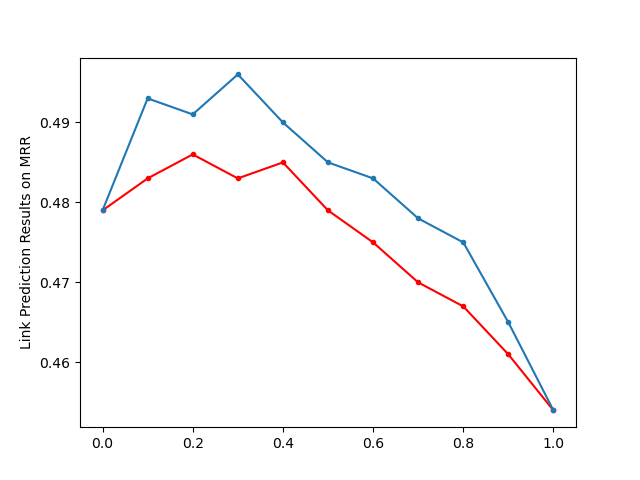}
  \includegraphics[width=0.6\linewidth]{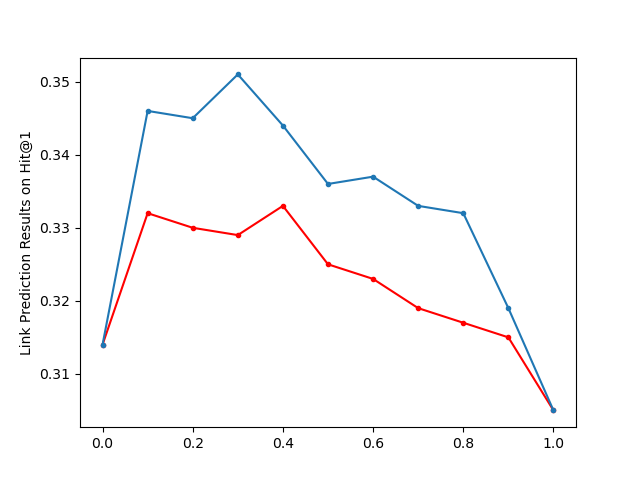}
  \caption{Impact of sampling proportion $\beta_1, \beta_2$ for two-stage (MANS-T, the \textcolor{red}{red} line) and hybrid (MANS-H, the \textcolor{blue}{blue} line) negative sampling. The experiments are based on FB15K dataset and TransE base score function.}
  \label{img:curve}
\end{figure}

It is worth mentioning that when $\beta_i=0.0 (i=1,2)$, both MANS-T and MANS-H degrade to normal negative sampling. When $\beta_i=1.0 (i=1,2)$, both of them become MANS-I. Thus, they can be baselines for comparison.
\par We could observe that the trends of MANS-T and MANS-H are almost identical. For MANS-T, the best proportion $\beta_1=0.3$, and for MANS-H the best proportion $\beta_2=0.4$. In the range of 0.1 to 0.4, MMKGE models trained with MANS-T and MANS-H perform better. Meanwhile, we could find that as the proportion of image negative sampling increases (when $\beta_1, \beta_2\ge 0.5$), the model performance would get down and might be worse than normal negative sampling.
In the range of 0.1 to 0.4, the performance of each strategy has just little changes most of the time. Therefore, the best choice for sampling proportion should most likely be in this range.

\subsection{RQ3: Adaptive Setting}
From the previous experiments, we could observe that the performance of MANS-A is close to and slightly better than MANS-H most of the time. In this section, we will dive into MANS-A and make further exploration to illustrate the rationality of its design and answer RQ3. 
\par
\begin{figure}[]
    \centering
    \includegraphics[width=0.8\linewidth]{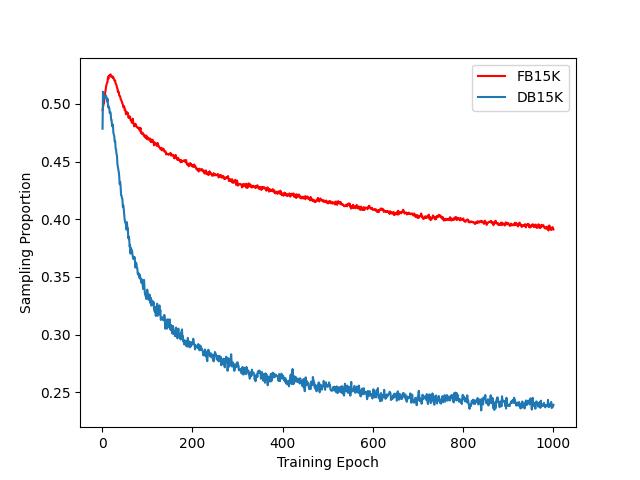}
    \caption{Trend of adaptive sampling proportion $\beta_3$ during whole training process on two datasets.}
    \label{img:adp}
\end{figure}

 We record the adaptive proportion $\beta_3$ for each batch of data in each training epoch and then calculate the average adaptive proportion of all the batches for each epoch. The trends of adaptive sampling proportion $\beta_3$  for different models and datasets in each training epoch are shown in Figure \ref{img:adp}.

\par  According to Figure \ref{img:adp}, the adaptive proportion $\beta_3$ usually becomes stable during the training process. We could pay attention to the stable part of each curve. Compared with the optimal proportions for MANS-H (which can be found in the previous section), we could find that design of MANS-A is reasonable as the adaptive proportion $\beta_3$ in MANS-A is close to the optimal settings in MANS-H.
For example, the stable sampling proportions on FB15K and DB15K are nearly 0.4 and 0.3. They are close to the optimal or sub-optimal $\beta_2$ of MANS-H. This suggests that the adaptive setting MANS-A would find the suitable proportion $\beta_3$ which is consistent with MANS-H but free of tuning. In summary, the design of MANS-A is reasonable and effective.

\subsection{RQ4: Efficiency}
As we mentioned earlier, MANS is more lightweight and efficient compared with existing methods because it is free of over-designed. Therefore, we evaluate the training speed of each NS method and list the results in Table \ref{tab:speed}, aiming to answer RQ4. The experiments are conducted on a single Nvidia GeForce RTX 3090 GPU.
\par From the table, we could find that the training speed of MANS is closer to the normal NS. Even the most complicated MANS-A is more efficient than several baselines. Though No-Samp \cite{DBLP:conf/www/LiJFGXCZ21} is very fast, it fails to perform well in MMKGE according to the link prediction results in Table \ref{tab:linkprediction}.
\par We also list the extra modules proposed by each method. Unlike random walks in SANS \cite{DBLP:conf/emnlp/AhrabianFSHB20} and entity clustering in EANS \cite{DBLP:journals/corr/abs-2210-06242}, our visual NS is not computationally intensive, which is the reason why MANS is lightweight enough. Besides, we have found in practice that NSCaching \cite{DBLP:conf/icde/ZhangYSC19} and No-Samp \cite{DBLP:conf/www/LiJFGXCZ21} would consume lots of memory and GPU resources, which is 1.13$\times$ (NSCaching \cite{DBLP:conf/icde/ZhangYSC19}) and 6.65$\times$ (No-Samp \cite{DBLP:conf/www/LiJFGXCZ21}) than MANS-A. In summary, MANS is lightweight and efficient enough and could make the training process faster compared with other NS methods. We have achieved a significant improvement in two tasks of KGC with our lightweight design.

\begin{table}[]
\centering
\caption{The training speed of different NS methods and the extra modules proposed by them}
\label{tab:speed}
\begin{tabular}{ccc}

\toprule
   Method       & Traning Speed(s/epoch) & Extra Module \\ \midrule
Normal \cite{DBLP:conf/nips/BordesUGWY13}    & 14.3                  &   -           \\
No-Samp \cite{DBLP:conf/www/LiJFGXCZ21}   & 0.2                  &   Full-batch Traning           \\
NSCaching \cite{DBLP:conf/icde/ZhangYSC19} & 16.7                  &   Entity Caching           \\
SANS  \cite{DBLP:conf/emnlp/AhrabianFSHB20}     &  16.6                &   Random Walks \\
EANS \cite{DBLP:journals/corr/abs-2210-06242}      &  60.9                &  Entity Clustering            \\ \midrule
MANS-I    &  14.8                &   \multirow{4}{*}{Visual NS}           \\
MANS-T    &  14.5                &              \\
MANS-H    &  15.1                &              \\
MANS-A    &  16.1                &     \\ \bottomrule

\end{tabular}
\end{table}

\subsection{RQ5: Embedding Visualization}
\begin{figure}[h]
  \centering
  \subfigure[Normal NS]{
  \includegraphics[width=0.45\linewidth]{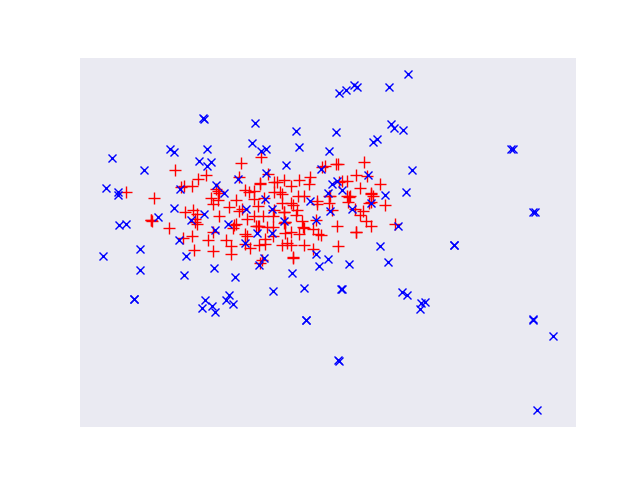}
  }
  \subfigure[MANS-T]{
  \includegraphics[width=0.45\linewidth]{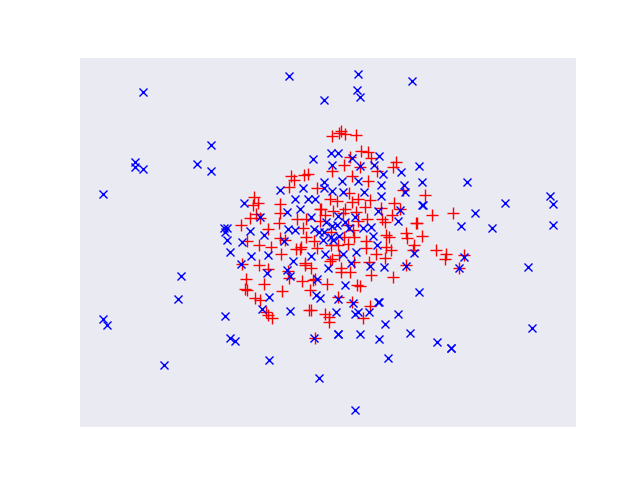}
  }
  \subfigure[MANS-H]{
  \includegraphics[width=0.45\linewidth]{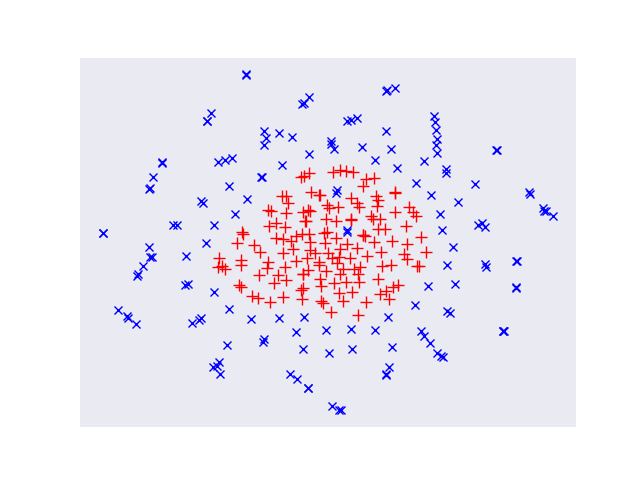}
  }
  \subfigure[MANS-A]{
  \includegraphics[width=0.45\linewidth]{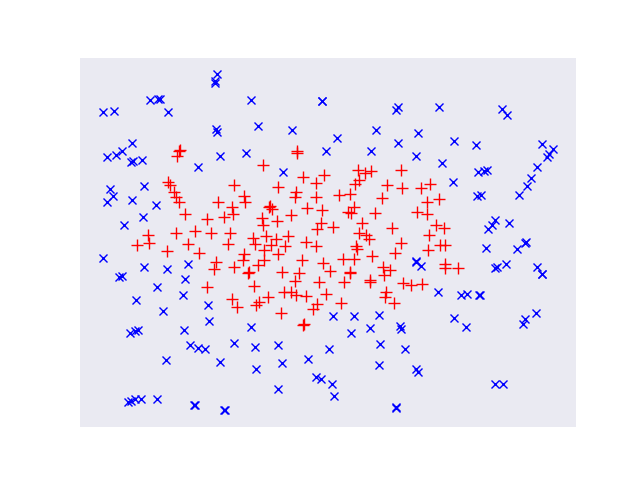}
  }
  \caption{Visualization using t-SNE for entity embeddings trained with different NS methods. The marker \textcolor{red}{$+$} and \textcolor{blue}{$\times$} represent the structural and visual embeddings.} 
  \label{img:vis}
\end{figure}
To evaluate the effectiveness of MANS and answer RQ5 in a straightforward view, we apply t-SNE to visualize the structural and visual embeddings of entities. We select the entities with type \textit{/award/award\_winner} in FB15K and the results are shown in Figure \ref{img:vis}.

\par From the visualization results, we could observe that the distribution of structural and visual embeddings of normal NS is close to each other. This means that the semantic information they express is relatively similar.
However, the embedding distribution of MANS-H and MANS-A shows a more clear boundary between the two kinds of embeddings compared with normal NS, which means the learned embeddings have more semantic information and the MMKGE model can clearly distinguish them to enhance the model performance, which is consistent with the link prediction performance in Table \ref{tab:linkprediction}. Thus, RQ5 is solved and we could conclude that MANS could guide the MMKGE model to learn meaningful and semantic-rich embeddings.
\section{Conclusion}
In this paper, we propose MANS, a modality-aware negative sampling method for MMKGE, which focuses on the alignment between different modal embeddings of a MMKGE model. MANS is the first NS method designed especially for MMKGE while achieving efficiency and effectiveness to solve the problems of existing NS methods. We first propose visual negative sampling (MANS-V) and extend MANS-V to three different settings called MANS-T, MANS-H, and MANS-A. Besides, we conduct comprehensive experiments on two public benchmarks and two classic tasks to demonstrate the performance of MANS compared with several state-of-the-art NS methods. In the future, we plan to conduct more in-depth research about MMKGE from two perspectives: (1) developing more robust solutions to achieve modal alignment and fusion of MMKG, (2) attempting to make co-design of the MMKGE model and NS method for better performance.
\section*{Acknowledgement}
This work is funded by Zhejiang Provincial Natural Science Foundation of China (No. LQ23F020017) and Yongjiang Talent Introduction Programme (No. 2022A-238-G).


\bibliographystyle{IEEEtran}
\bibliography{ref}

\end{document}